\documentclass[11pt]{article}

\usepackage[final]{acl}

\usepackage{times}
\usepackage{latexsym}

\usepackage[T1]{fontenc}

\usepackage[utf8]{inputenc}

\usepackage{microtype}

\usepackage{inconsolata}

\usepackage{graphicx}
\usepackage{threeparttable}
\usepackage{booktabs}
\usepackage{amsmath}
\usepackage{amssymb}
\usepackage{multirow}
\usepackage[caption=false]{subfig}
\usepackage{enumitem}
\usepackage{makecell}
\usepackage{wrapfig}
\usepackage{framed}

\title{Enhancing Online Recruitment with Category-Aware MoE and LLM-based Data Augmentation}

\author{
  Minping Chen\textsuperscript{1, }\footnotemark[1], Bing Xu\textsuperscript{3, }\footnotemark[1],
  Zulong Chen\textsuperscript{3, }\footnotemark[2], \\
  \textbf{Chuanfei Xu\textsuperscript{4, }\footnotemark[2], 
  Ying Zhou\textsuperscript{5}, Zui Tao\textsuperscript{6}, Zeyi Wen\textsuperscript{1, 2, }\footnotemark[2]} \\
  \textsuperscript{1}HKUST (GZ)\quad
  \textsuperscript{2}HKUST\quad
  \textsuperscript{3}Alibaba Group \\
  \textsuperscript{4}Guangdong Laboratory of Artificial
Intelligence and Digital Economy (SZ)\\
\textsuperscript{5}Zhijiang Lab\quad
\textsuperscript{6}The Hong Kong Polytechnic
University\\
  \small{
   \textbf{Correspondence:} \texttt{zulong.czl@alibaba-inc.com}, \texttt{xuchuanfei@gml.ac.cn}, \texttt{wenzeyi@hkust-gz.edu.cn}
 }
 \\
}

\begin{document}
\maketitle

{\renewcommand{\thefootnote}{}
\footnotetext[0]{\textsuperscript{*}~Equal contribution.}
\footnotetext[0]{\textsuperscript{\dag}~Corresponding author.}}

\begin{abstract}
Person-Job Fit (PJF) is a critical component for online recruitment. Existing approaches face several challenges, particularly in handling low-quality job descriptions and similar candidate-job pairs, which impair model performance. To address these challenges, this paper proposes a large language model (LLM) based method with two novel techniques: (1) LLM-based data augmentation, which polishes and rewrites low-quality job descriptions by leveraging chain-of-thought (COT) prompts, and (2) category-aware Mixture of Experts (MoE) that assists in identifying similar candidate-job pairs. This MoE module incorporates category embeddings to dynamically assign weights to the experts and learns more distinguishable patterns for similar candidate-job pairs. We perform offline evaluations and online A/B tests on our recruitment platform. Our method relatively surpasses existing methods by 2.40\% in AUC and 7.46\% in GAUC, and boosts click-through conversion rate (CTCVR) by 19.4\% in online tests, saving millions of CNY in external headhunting expenses.
\end{abstract}

\section{Introduction}
The swift advancement of Internet technology has transformed online recruitment into a widely used web service for job seekers and recruiters~\cite{geyik2018talent,kenthapadi2017personalized,paparrizos2011machine}. Due to the significant increase in usage of online recruitment platforms, Person-Job Fit (PJF)~\cite{qin2023comprehensive} has emerged as an effective solution to automatically measure the matching degree between a job and a candidate. 


In recent years, the PJF task has been regarded as a text matching task, by exploiting the rich semantic information in resumes and job descriptions (JDs). Various neural networks have been applied to enhance the encoding of the resumes and JDs, including CNN~\cite{zhu2018person,he2021finn,zhenhong2021person}, RNN~\cite{qin2018enhancing,qin2020enhanced,yan2019interview} and GNN~\cite{wang2022person} et al. Along with improving the text representation, existing methods also explore the candidate-job interaction. Such efforts include employing historically accepted and rejected applications~\cite{le2019towards}, capturing users’ dynamic preferences from their multi-behavioral sequences such as click, invite/apply, chat~\cite{yang2022modeling} or searching histories~\cite{hou2022leveraging}, modeling the two-way interaction~\cite{yang2022modeling,zheng2023reciprocal}, and multi-stage interaction~\cite{zheng2024mirror}. 

Despite the notable progress, existing methods still encounter these challenges: (i) Numerous low-quality job descriptions (JDs) exist in our online recruitment system (about 25\% training samples), i.e., with short content and poor information. JD is a crucial input for the PJF task, and poor JDs hinder the model performance. However, this issue remains unaddressed by existing methods. ii) While existing methods have explored multiple interaction strategies, they struggle with managing similar candidate-job pairs, i.e., the job requirements and the work experience of the candidates are similar, but they differ in some key aspects. For instance, candidates for a data analyst position may share relevant experience with algorithm engineers, e.g., basic data processing and algorithm development. However, algorithm engineers primarily design and optimize algorithms for applications, while data analysts interpret data to provide insights for business decisions. Therefore, a data analyst candidate is not a suitable match for an algorithm engineer position.

To address these challenges, we propose a large language model (LLM) based method with data augmentation and mixture of experts (MoE)~\cite{jacobs1991adaptive,shazeer2017outrageously,lepikhin2020gshard} architecture. Specifically, we introduce an LLM-based data augmentation module to polish or rewrite the low-quality JDs. Chain-of-thought (COT)~\cite{wei2022chain} prompt templates are carefully designed to instruct the LLM to complete this task following the given requirements, with several resumes that have passed the interviews as references. Furthermore, to improve the model's ability to identify similar candidate-job pairs, we propose a category-aware MoE module that leverages the category information of the jobs and candidates. This category information is input to the gating net to dynamically assign suitable weights for each expert. Each expert, acting like a human expert, is responsible for processing a different scenario and capturing the fine-grained distinction of similar texts and historical interactions. We summarize the contributions of this paper as follows. 
\begin{enumerate}
    \item We introduce an LLM-based data augmentation strategy to refine low-quality job descriptions, a challenge that remains unaddressed in existing methods, thereby improving the overall model performance.
    \item We propose a novel category-aware MoE module to enhance the matching degree assessment. With this module, our method can learn more distinguishable patterns for similar candidate-job pairs.
    \item We conduct offline experiments and online A/B tests on our recruitment platform. Our method relatively outperforms the state-of-the-art method by 2.40\% in AUC and 7.46\% in GAUC, and boosts the click-through conversion rate (CTCVR) in online tests by 19.4\%. 
\end{enumerate}

\begin{figure*}[t]
\begin{center}
    \includegraphics[width=0.95\linewidth]{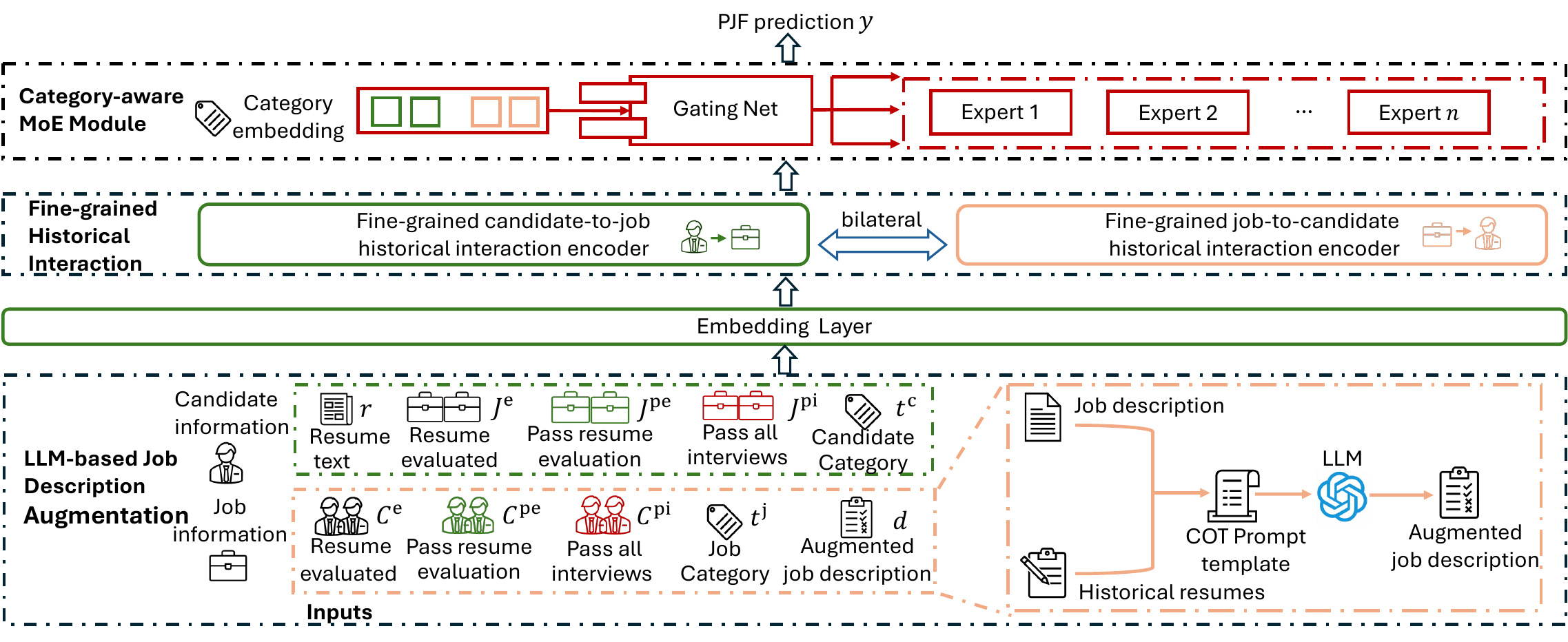}
\end{center}
\caption{The overview of our method. } 
\label{model_overview}
\end{figure*}


\section{Methodology}
\subsection{Notation and Problem Statement}
\label{section_notation}
Let $C=\{c_1, c_2,...,c_m\}$ be the candidates, and $J=\{j_1, j_2,...,j_n\}$ be the jobs, with $m$ and $n$ as their sizes. Each candidate $c \in C$ has a resume text $r$ and a category $t^{\rm c}$. Each job $j \in J$ has a description text $d$ and a category $t^{\rm j}$. Candidates have historical job interaction sequences: jobs that the resume of the candidate has been evaluated $\mathcal{J}^{\rm e} = \{j_1, j_2,...j_{k_1}\}$, jobs passed resume evaluation $\mathcal{J}^{\rm pe} = \{j_1, j_2,...j_{k_2}\}$, and jobs passed the interviews $\mathcal{J}^{\rm pi} = \{j_1, j_2,...j_{k_3}\}$. Similarly, jobs have candidate interaction sequences: evaluated candidates $\mathcal{C}^{\rm e} = \{c_1, c_2,...c_{k_4}\}$, candidates passed resume evaluation $\mathcal{C}^{\rm pe} = \{c_1, c_2,...c_{k_5}\}$, and candidates passed the interviews $\mathcal{C}^{\rm pi} = \{c_1, c_2,...c_{k_6}\}$. The numbers $\{k_1, k_2,...k_6\}$ represent the sequence sizes. Given these inputs, our model predicts if candidate $c_i$ matches job $j_i$ with binary output $y_i$.

\subsection{Overview of Our Method}
We present an overview of our method, as shown in Figure~\ref{model_overview}. It comprises three modules: LLM-based job description augmentation module, fine-grained historical interaction module and category-aware MoE module. The model processes the candidate information and job information, converting them into embeddings. To tackle the issue of low-quality job descriptions (JDs), we introduce a data augmentation strategy based on LLMs. Then the fine-grained historical interaction module learns the bilateral and comprehensive interactions between the candidates and jobs, aligning with the recruitment process to enhance effectiveness. The category-aware MoE module further refines the interaction encoding by incorporating category information to address the challenge of similar candidate-job pairs, ultimately producing predictions.

\subsection{LLM-based JD Augmentation} 
\label{JD_module}
Our online recruitment system has many short, low-quality job descriptions (25\% of training samples). To address this, we leverage an LLM to enhance JDs below a certain character-level length threshold $l$, as updating all JDs may introduce noise. We conduct multiple rounds of optimization on a JD subset to enhance the COT prompt template. The prompt is based on the original JD and its historical resumes from candidates who passed the interviews, offering references for the LLM to refine the JDs. To tackle potential LLM hallucination issues, we carefully design the prompt, as illustrated in Figure~\ref{prompt_design}. The \textbf{system instruction} guides the LLM on its role and introduces the task. The LLM needs to act as an "HR expert" with a professional background in human resources of various industries, and tasks like resume analysis and job matching.

In the \textbf{requirement part}, we provide clear task descriptions, which is divided into multiple steps based on the COT technique. First, the LLM evaluates the completeness of the original JD, then extracts the key information, e.g., job position and professional skills. For JDs that lack historically matched resumes, the LLM leverages its comprehensive HR expertise to enhance them. Otherwise, the LLM summarizes the common industry areas and professional requirements in multiple resumes, and integrates the results into the original JD to improve it. To ensure quality, we require the LLM to retain over 70\% of the original keywords, avoid complexity or vagueness, and maintain clear, concise professionalism. These guidelines help produce effective JDs and prevent excessive rewriting noise. The full prompt is presented in Appendix~\ref{prompt_example}.


\subsection{Fine-grained Historical Interaction} 
\begin{figure}[t]
\begin{center}
    \includegraphics[width=\linewidth]{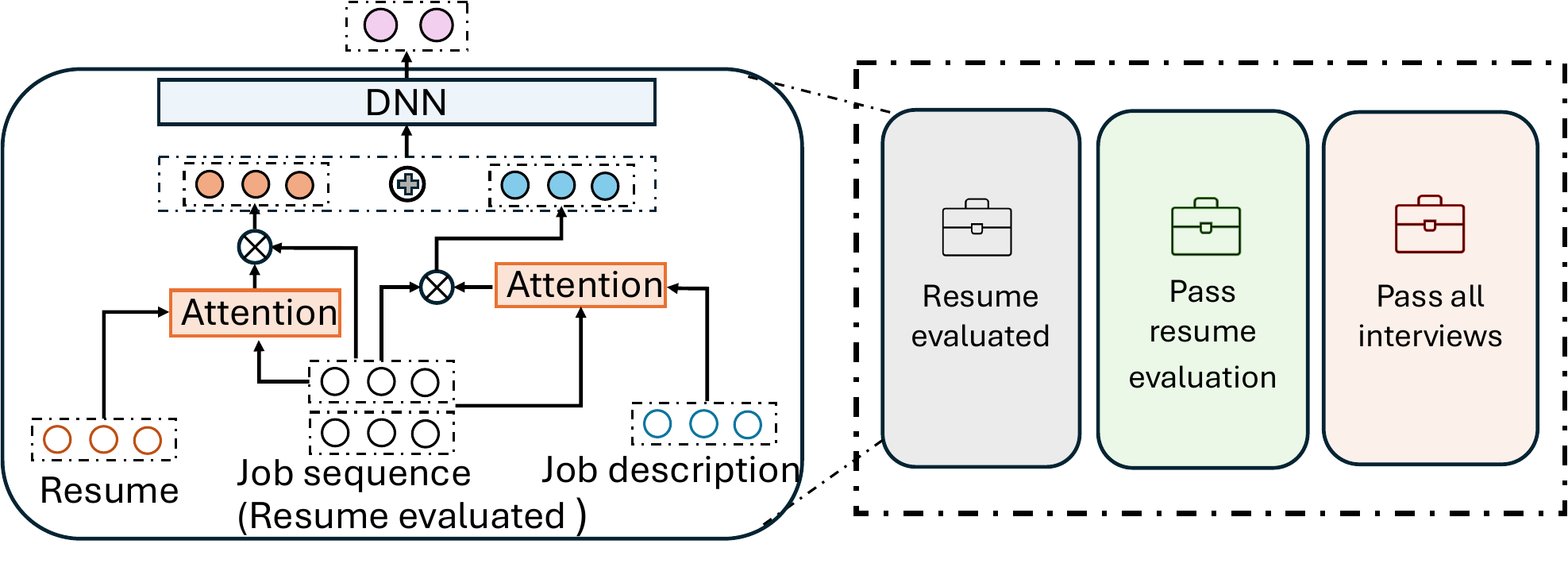}
\end{center}
\caption{The architecture of the fine-grained candidate-to-job historical interaction encoder.} 
\label{interaction_module}
\end{figure}

The fine-grained historical interaction module uses two identical encoders to learn the bilateral interaction between candidates and jobs. The only difference between the encoders is their input. We present one of them in Figure~\ref{interaction_module}. For simplicity, we use $\{r_i, d_i, \mathcal{J}_{i}^{\rm e}, \mathcal{J}_{i}^{\rm pe}, \mathcal{J}_{i}^{\rm pi}, \mathcal{C}_{i}^{\rm e}, \mathcal{C}_{i}^{\rm pe}, \mathcal{C}_{i}^{\rm pi}\}$ to represent the input embeddings processed by the interaction encoder. Our bilateral interaction is more fine-grained and comprehensive than existing methods. We break down interaction modeling into procedures reflecting the recruitment process: finishing resume evaluation, passing resume evaluation and passing all interviews. For each procedure, we both learn internal and external interactions. Internal interaction involves the JD embedding $d_i$ and its historical interacted candidate sequences $\mathcal{C}_{i}^{\rm e}/\mathcal{C}_{i}^{\rm pe}/ \mathcal{C}_{i}^{\rm pi}$ (or the resume embedding $r_i$ and its historical interacted job sequences $\mathcal{J}_{i}^{\rm e}/\mathcal{J}_{i}^{\rm pe}/\mathcal{J}_{i}^{\rm pi}$), and external interaction involves $d_i$ and $\mathcal{J}_{i}^{\rm e}/\mathcal{J}_{i}^{\rm pe}/\mathcal{J}_{i}^{\rm pi}$ (or $r_i$ and $\mathcal{C}_{i}^{\rm e}/\mathcal{C}_{i}^{\rm pe}/ \mathcal{C}_{i}^{\rm pi}$). We utilize the multi-head attention~\cite{vaswani2017attention} to learn these interactions, and take the internal candidate-to-job interaction of finishing resume evaluation as an example:
\begin{eqnarray*}
\resizebox{0.9\linewidth}{!}{$
\mathrm{Attention}(Q,K,V) = \mathrm{softmax}(\frac{QK^T}{\sqrt{d_{\rm k}}})V,$}\\
\label{att}
\resizebox{\linewidth}{!}{$
\begin{split}
    \mathrm{MultiHead}(r_i,\mathcal{J}_{i}^{\rm e}) = \mathrm{Concat}(head_{i}, head_{2},...head_{h})W^{\rm O}, \\
\mathrm{where}~head_{i}=\mathrm{Attention}(r_{i}W^{\rm Q}_{i},\mathcal{J}_{i}^{\rm e}W^{\rm K}_{i},\mathcal{J}_{i}^{\rm e}W^{\rm V}_{i})
\end{split}$}
\label{multihead_att}
\end{eqnarray*}
Here $W^{\rm Q}_{i}\in \mathbb{R}^{d_{\rm model}\times d_{\rm k}}$, $W^{\rm K}_{i}\in \mathbb{R}^{d_{\rm model}\times d_{\rm k}}$, $W^{\rm V}_{i}\in \mathbb{R}^{d_{\rm model}\times d_{\rm v}}$,  $W^{\rm O}\in \mathbb{R}^{hd_{\rm v}\times d_{\rm model}}$ are the weight matrices, $d_{\rm model}$ is dimension of the embedding, $h$ is the number of attention heads and $d_{\rm k} = d_{\rm v} = d_{\rm model} / h$.  For the external interaction in this example, we only need to replace $Q$ with $d_iW^{\rm Q}_{i}$. The learned bilateral interactions are concatenated into a DNN for feature fusion. 


\subsection{Category-aware MoE Module} 

Initially, a category vocabulary is defined by the HRs in our company based on industry standards, and the reliability of category assignment for the jobs and candidates is guaranteed through manual verification. Subsequently, we learn an embedding matrix $E_{\rm c}\in \mathbb{R}^{n_{\rm c} \times d_{\rm e}}$ for this vocabulary, where $n_{\rm c}$ is the vocabulary size and $d_{\rm e}$ is the embedding dimension. $E_{\rm c}$ is randomly initialized and optimized during training. Each expert in the category-aware MoE module is a feed-forward network (FFN), aiming to learn distinct candidate-job matching patterns across various job categories. Like a human expert, each expert specializes in different scenarios that cover multiple job categories. Furthermore, they capture the distinction and learn a feature fusion of the bilateral and fine-grained candidate-job interactions. In other words, the MoE module leverages not only the category information but also the representation and interaction distinctions to identify similar candidate-job pairs. The gating net is a two-layer FFN, guided by the category embedding $e_{\rm c}$ to dynamically assign appropriate weights to each expert. The final prediction is based on the weighted sum of the output of each expert:
\begin{equation*}
G_i = \mathrm{softmax}(W_2^{\rm G}(\mathrm{ReLU}(W_1^{\rm G}e_{\rm c} + b_1^{\rm G})) + b_2^{\rm G}),    
\end{equation*}
\begin{equation*}
\resizebox{\linewidth}{!}{$
    E_i(x) = W_3^{\rm E}(\mathrm{ReLU}(W_2^{\rm E}(\mathrm{ReLU}(W_1^{\rm E}x + b_1^{\rm E}) + b_2^{\rm E})) + b_3^{\rm E},$}
\end{equation*}
\begin{equation*}
y=\sum_{i=1}^{n_{\rm e}}G_iE_i(x),
\end{equation*}
where $E_i(x)$ is the output of the $i$-th expert, $G_i$ is the gating weight of the $i$-th expert, $x$ is the joint representation of the input, $n_{\rm e}$ is the number of experts, $W_i^{\rm G}$ and $W_i^{\rm E}$ are the weights of the gating network and expert, and $y$ is the final prediction. 

\begin{figure}[t]
\begin{center}
    \includegraphics[width=0.95\linewidth]{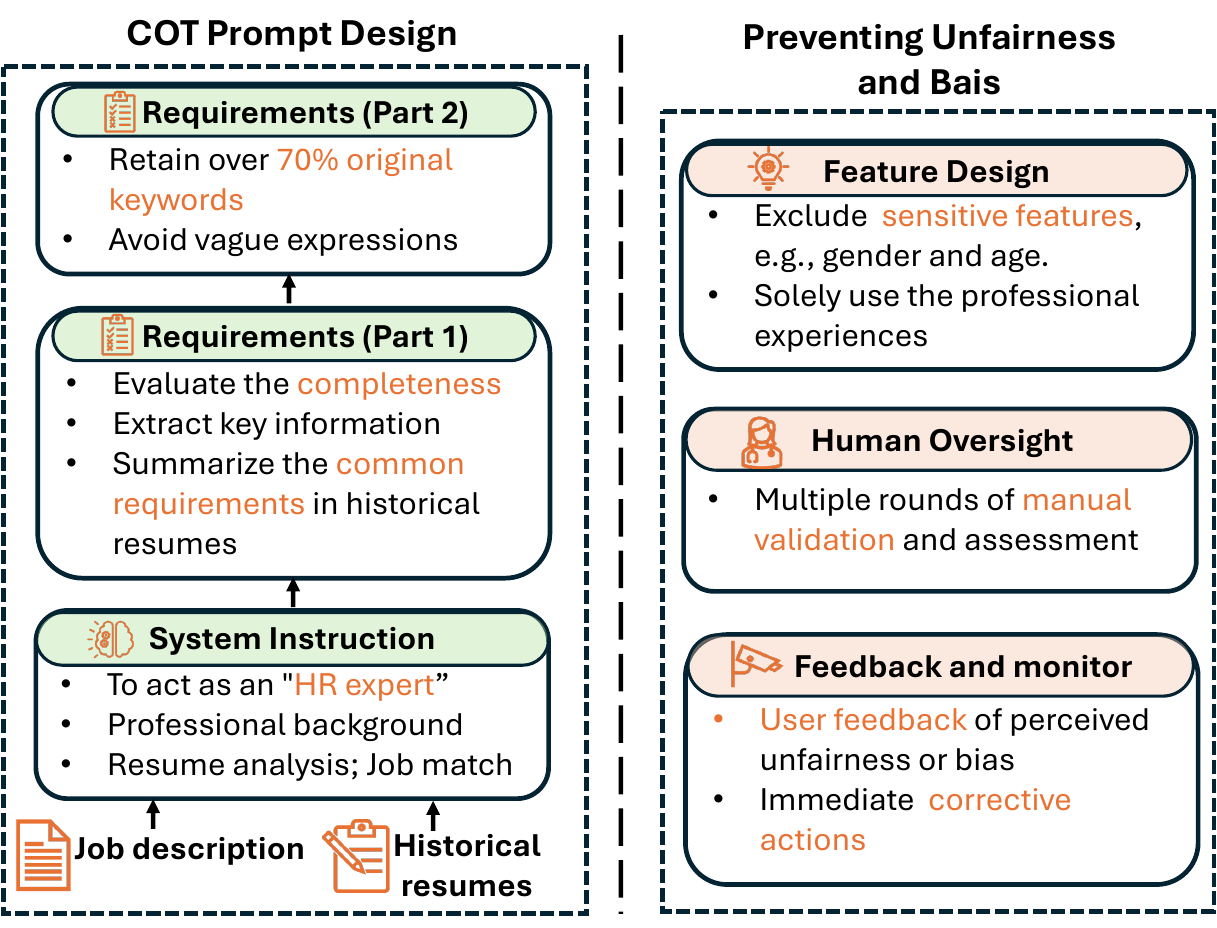}
\end{center}
\caption{Prompt and strategies design in our method.} 
\label{prompt_design}
\end{figure}

We train our model in a pair-wise way using the BPR loss~\cite{rendle2012bpr} to optimize the ranking between positive and negative samples, adding a regularization item to avoid overfitting:
\begin{equation*}
\textbf{\resizebox{\linewidth}{!}{$
            \displaystyle
\begin{aligned}
\mathcal{L} = -\frac{1}{|B|} \sum_{(y_{i}^{+},y_{i}^{-})\in B} \mathrm{log}(\sigma(y_{i}^{+} - y_{i}^{-})) +
\lambda (\frac{1}{|B|} \sum_{(y_{i}^{+},y_{i}^{-})\in B} (y_{i}^{+})^2+(y_{i}^{-})^2),
\end{aligned}
$}}
\label{loss_fuction}
\end{equation*}
where $B$ is a batch of training data, $y_{i}^{+}$ and $y_{i}^{-}$ are the prediction scores of the positive and negative samples, respectively, $\sigma$ denotes the sigmoid function, and $\lambda$ is the regularization weight. 

\subsection{Strategies for Preventing Bias}
\label{strategies_for_bias}
To ensure that the AI model in our system does not introduce unfairness or biases to the decision-making, we develop several strategies in Figure~\ref{prompt_design}:

\textbf{Feature design for fairness}: Our model intentionally excludes sensitive features that may lead to unfairness and biases as inputs, e.g., gender and age. However, neural networks are susceptible to bias arising from proxy features, such as educational institutions, graduation years, and geographic location. To address this, we do not use such proxy features as explicit inputs. We solely use the professional experiences of the talents for PJF modeling, ensuring that the model focuses on evaluating professional qualifications, skills, and experiences rather than features having a risk of introducing unfairness and biases. 

\textbf{Human Oversight and Feedback}: It is important to note that our method is used as a decision support tool rather than an automated filter. Candidates are ranked at the user interface based on model predictions, and recruiters retain full control over hiring decisions. Our recruitment process is fortified by rigorous human oversight, with multiple rounds of manual validation and assessment to ensure fairness. Additionally, it is equipped with a feedback and monitoring mechanism, empowering users to report instances of perceived bias. This facilitates immediate corrective actions and reinforces our commitment to fairness. 

\section{Experiments}
\subsection{Data and Baselines}
\paragraph{Data Collection} In the PJF problem, the only one open-source dataset, i.e., Aliyun dataset (which has 36274 training samples, 300 validation samples, and 300 test samples )\footnote{https://tianchi.aliyun.com/competition/entrance/231728} is not suitable for our scenario, as it misses some key information (e.g., job categories and historical interactions). Thus, we collect a real-world dataset from our online recruitment system, creating training data by randomly sampling matched candidate-job pairs for positive samples and unmatched candidates for negative samples. The training data contains about 8 million samples (47K jobs and 0.8 million resumes). The test data is also collected from our system, consisting of 25K samples (2K jobs and 20K resumes). Note that our data split is strictly temporal: training data is collected from an earlier period (up to mid-July 2024), while test data is from a later period (from mid to end of July 2024), ensuring no future information leaks into training. Additionally, we perform candidate and job deduplication based on their unique IDs. Our code is available here\footnote{https://github.com/Chan-1996/LLM-PJF}. 

\paragraph{Evaluation Metrics}
\label{metric}
For offline evaluations, we use AUC, grouped AUC (GAUC), normalized discounted cumulative gain (NDCG), and average precision score (AP) as evaluation metrics. AUC and GAUC are not fully linearly correlated because GAUC groups samples by jobs, and higher GAUC is achieved only when samples within the same job are ranked more accurately. For online A/B tests, we use click-through conversion rate (CTCVR)~\cite{ma2018entire} as the evaluation metric. A higher CTCVR indicates recruiters find suitable candidates more accurately and with less effort, reducing the number of resumes to review. Detailed definitions of these metrics are in Appendix~\ref{setup}.




\paragraph{Baselines}
We compare our method with various baselines: Logistic Regression (LR), XGBoost~\cite{chen2016xgboost}, DSSM~\cite{huang2013learning}, BGE-Raw~\cite{chen2024bge}, PJFNN~\cite{zhu2018person}, IPJF~\cite{le2019towards}, PJFFF~\cite{jiang2020learning}, SHPJF~\cite{hou2022leveraging}, and CONFIT~\cite{yu2024confit}. Introduction of these baselines and other implementation details are presented in Appendix~\ref{setup}.


\begin{table}[t]
\centering
\caption{Performance comparison on our dataset. } 
\resizebox{\linewidth}{!}{
\newcommand{\tabincell}[2]{\begin{tabular}{@{}#1@{}}#2\end{tabular}}
\begin{tabular}{l|l|cccc}
\toprule
~ & Method & AUC & GAUC & NDCG & AP \\
\midrule
\midrule
\multirow{2}*{Traditional ML} & LR & 0.672 & 0.615 & 0.827 & 0.235 \\ 
~ & XGBoost & 0.673 & 0.616 & 0.830 & 0.237 \\ 
\midrule
\multirow{2}*{Similarity based} & DSSM & 0.653 & 0.609 & 0.819 & 0.216\\ 
~ & BGE-Raw & 0.606 & 0.592 & 0.803 & 0.187 \\ 
\midrule
\multirow{6}*{PJF model} & PJFNN & 0.635 & 0.634 & 0.811 & 0.200 \\ 
~ & IPJF & 0.669 & 0.630 & 0.832 & 0.237  \\ 
~ & PJFFF & 0.650 & 0.643 & 0.804 & 0.197\\
~ & SHPJF & 0.651 & 0.630 & 0.810 & 0.204\\
~& CONFIT & 0.707 & 0.661 & \textbf{0.842} & 0.260 \\
\cmidrule{2-6}
~ & Ours & \textbf{0.724} & \textbf{0.709} & 0.837 & \textbf{0.262} \\
\bottomrule
\end{tabular}}
\label{overall_pefromance}
\end{table}

\subsection{Main Results}
\paragraph{Offline Evaluation} The offline comparison is presented in Table~\ref{overall_pefromance}. First, LR and XGBoost can effectively rank obvious positive samples, and get high NDCG due to their sensitivity to distinct features like exact matches. They also achieve better performance than the similarity-based deep learning methods. This indicates that cosine similarity cannot well distinguish different matching degrees in the PJF task. Compared with CONFIT, our method achieves a relative improvement of 2.40\% and 7.46\% in AUC and GAUC, respectively. Although the NDCG of our method drops, the degradation is negligible (i.e., only -0.59\%). This is because our solution works better for matching than ranking (higher AUC), while CONFIT is a recall model, focusing on ranking (slight higher NDCG). However, we achieve better overall performance.

\begin{table}[t]
\centering
\caption{Performance comparison on Aliyun dataset.}
\centering
\resizebox{0.78\linewidth}{!}{
\begin{threeparttable}
\begin{tabular}{c|cccc}
\toprule
Method & AUC & GAUC & NDCG & AP \\
\midrule
CONFIT & 0.523 & 0.494 & 0.814 & 0.494 \\
Ours & \textbf{0.558} & \textbf{0.565} & \textbf{0.884} & \textbf{0.674} \\
\bottomrule
\end{tabular}
\end{threeparttable}}
\label{aliyun_results}
\end{table}

We also evaluate on the Aliyun dataset to verify the effectiveness of our method when key information like job categories and historical interactions is not available. As shown in Table~\ref{aliyun_results}, our method consistently outperforms CONFIT in all metrics. This suggests that better feature modeling also contributes to our performance improvement.

\begin{wrapfigure}[8]{r}{0.2\textwidth}
    \centering
     \hspace{-0.05\textwidth} 
    \includegraphics[width=1.0\linewidth]{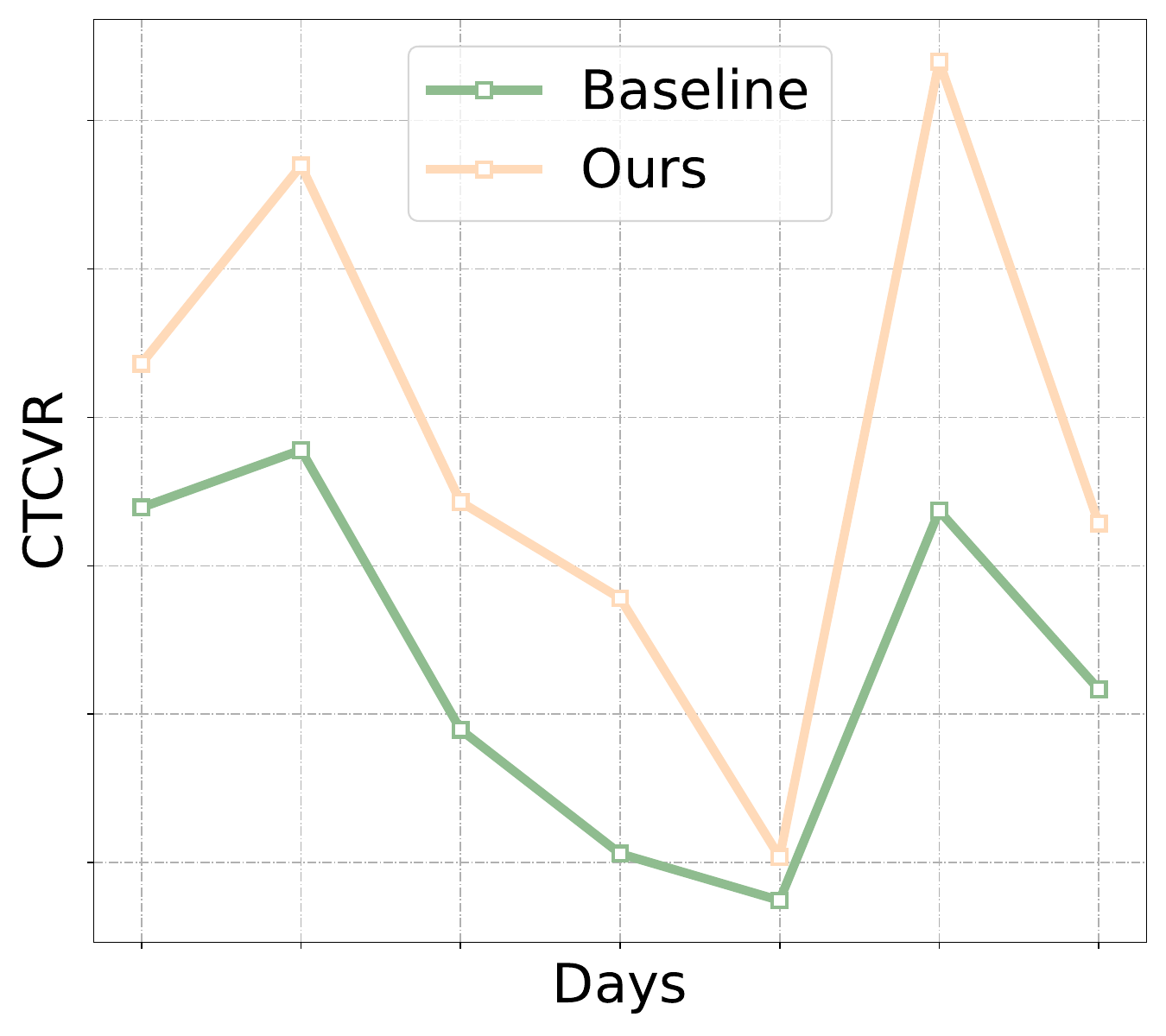}
    \caption{A/B tests.}
    \label{ab_test_result}
\end{wrapfigure}

\paragraph{Online Evaluation} We perform online A/B tests for seven working days on our recruitment platform, strictly following the principles of Google, e.g., randomization and single key metric. The results are shown in Figure~\ref{ab_test_result}. The online baseline is PJFFF, while CONFIT isn't compared due to its high overall latency (500ms vs. the required <300ms). For fair comparison, the job seekers and recruiters for the two methods were randomly selected in a 1:1 ratio. Additionally, our A/A test shows no significant difference between the groups, i.e., p-value $> 0.05$. Our method outperforms 
\setlength{\abovecaptionskip}{0.3pt} 
the baseline for all days. Due to confidentiality, we do not disclose the actual CTCVR values, but our method shows an average relative improvement of 19.38\%, statistically significant with a p-value of $0.0035<0.01$. Beyond CTCVR, we also monitor daily AUC metric, real-time user feedback through likes and dislikes, collection of bad cases, and regular user surveys. Moreover, our method helps save millions of CNY in headhunting expenses, as nearly 40\% of hires come from our internal talent database, reducing dependence on external channels. In terms of system efficiency, the LLM data augmentation is processed offline to ensure the overall latency remains below 300ms, and our model can perform inference even on CPUs since it is only about 0.5B. Besides, our system updates in real-time with a T+1 mode, where JDs are updated the following day, allowing new JDs with emerging skills to enter the pipeline.

\begin{table}[t]
\centering
\caption{Ablation study.} 
\resizebox{0.85\linewidth}{!}{
\newcommand{\tabincell}[2]{\begin{tabular}{@{}#1@{}}#2\end{tabular}}
\begin{tabular}{l|cc}
\toprule
Method & AUC &  GAUC  \\
\midrule
\midrule
Ours (full model) & \textbf{0.724}  & \textbf{0.709} \\
w/o JD Aug. & 0.718 & 0.702  \\
w/o MoE & 0.679 & 0.665 \\
w/o Int. & 0.717 & 0.695  \\
w/o JD Aug. \& MoE & 0.675 & 0.664 \\
w/o JD Aug. \& MoE  \& Int. & 0.664 & 0.648 \\
\bottomrule
\end{tabular} 
}
\label{ablation_study}
\end{table}

\subsection{Discussion}
\label{sec_ablation_study}
\paragraph{Ablation Study} We disable each module in our method separately to verify the effectiveness of them, as shown in Table~\ref{ablation_study}. \textit{w/o JD Aug.} and \textit{w/o MoE} refer to the model without using the LLM-based JD augmentation and the category-aware MoE module, respectively. \textit{w/o Int.} excludes the fine-grained historical interaction encoding, using only the interaction of passing the resume evaluation. Disabling any module or their combinations leads to performance drops, confirming their effectiveness for the PJF task. The most significant performance drop occurs when the category-aware MoE module is removed, highlighting its crucial role in enhancing the model's ability to differentiate similar candidate-job pairs.

\paragraph{Impact of the JD Augmentation} To evaluate how the length threshold $l$ affects performance, we evaluate our method with various $l$, as shown in the first seven rows of Table~\ref{threshold_exp}. Methods using JD augmentation consistently outperform the one without it, labeled \textit{Ours-w/o JD Aug.}, proving its effectiveness. Moreover, $l$ has small effect on performance, highlighting the robustness of our strategy. To assess whether the data augmentation enhances model performance on samples with low-quality JDs, we calculate the AUC and GAUC for these test samples, which comprise 9\% of the test set. The last three rows in Table~\ref{threshold_exp} show that the model with JD augmentation significantly outperforms the one without by 3.4\% in AUC and 7.0\% in GAUC. We also test one of our baselines, PJFFF, when provided with our augmented JDs, observing notable improvements over the vanilla PJFFF. This further supports the effectiveness of our LLM-based JD augmentation strategy.

\begin{table}[t]
\centering
\caption{Impact of the job description augmentation.} 
\resizebox{\linewidth}{!}{
\newcommand{\tabincell}[2]{\begin{tabular}{@{}#1@{}}#2\end{tabular}}
\begin{tabular}{l|c|c}
\toprule
Threshold $l$ & AUC (all) &  GAUC (all)  \\
\midrule
\midrule
$l$ = 200 & 0.724 & \textbf{0.709} \\
$l$ = 300 & 0.726 & 0.705 \\
$l$ = 350  & 0.726 & 0.706 \\
$l$ = 400 & 0.726 & 0.706 \\
w/o using $l$ & 0.724 & 0.704 \\
\midrule
Ours-w/o JD Aug. & 0.718 & 0.702 \\
Ours & \textbf{0.724} & \textbf{0.709} \\
\midrule
PJFFF & 0.650 & 0.643 \\
PJFFF + JD Aug. & \textbf{0.665} & \textbf{0.652} \\
\midrule
\midrule
Method & AUC (JD len $<200$) & GAUC (JD len $<200$)\\
\midrule
Ours-w/o JD Aug. & 0.698 & 0.512   \\
Ours &  \textbf{0.732} & \textbf{0.582}\\
\bottomrule
\end{tabular}}
\label{threshold_exp}
\end{table}

\paragraph{Impact of the LLM Hallucination Issue} As discussed in Section~\ref{JD_module}, we carefully design the prompt template to reduce LLM hallucination. To assess its effectiveness, we randomly sample fifty LLM-augmented JDs covering eight job categories, and manually check them for hallucination issues. The results are shown in Table~\ref{hallucination}. Only one JD suffers from the hallucination issue, and the ratio is 2\%, indicating the minor influence of the hallucination issue. Moreover, our system incorporates human oversight as introduced in Section~\ref{strategies_for_bias} to further minimize the effects of hallucination issues. 

\begin{table}[h]
\centering
\caption{Statistics on hallucination issues.}
\centering
\resizebox{\linewidth}{!}{
\begin{threeparttable}
\begin{tabular}{c|ccc}
\toprule
\#No Hallucination & \#Minor\footnotemark[1]  &  \#Moderate\footnotemark[2]  & \#Severe\footnotemark[3]   \\
\midrule
\midrule
49 & 1 & 0 & 0 \\
\bottomrule
\end{tabular}
\begin{tablenotes}
    \small
    \item \textsuperscript{1}Minor: Slight variations. \textsuperscript{2}Moderate: Noticeable differences. \textsuperscript{3}Severe: Significant deviation.
   \end{tablenotes}
\end{threeparttable}}
\label{hallucination}
\end{table}


\begin{figure}[b]
\begin{center}
    \includegraphics[width=0.95\linewidth]{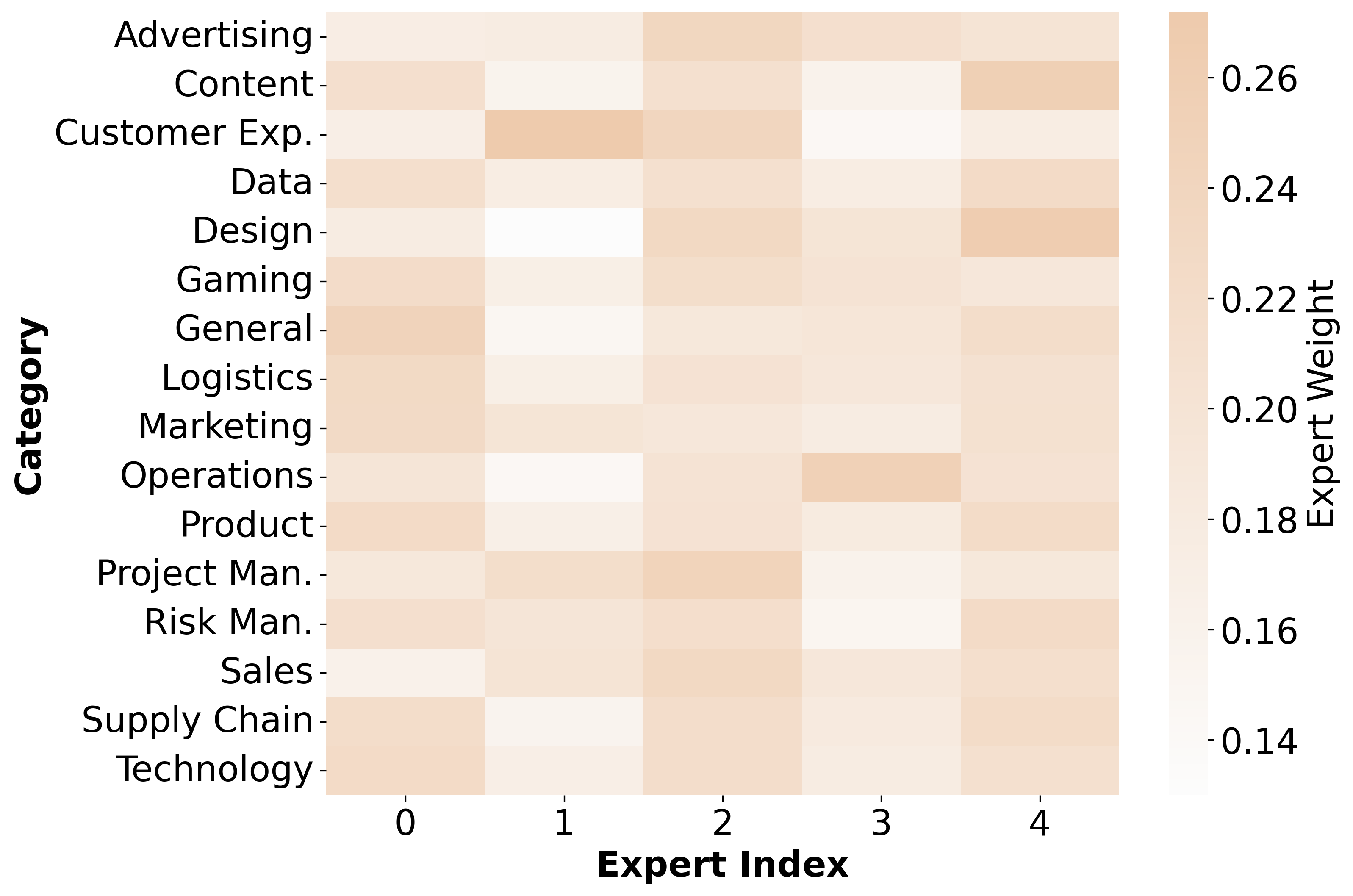}
\end{center}
\caption{Visualization of the expert weights.} 
\label{visualization}
\end{figure}

\paragraph{Expert Weights Visualization} Figure~\ref{visualization} shows the expert weights for various job categories. Specifically, we analyze the test set, calculating the average expert weight for each job category. Contributions of the five experts vary significantly across different person-job fitting scenarios, each encompassing several job categories. This highlights our method's ability to adaptively integrate the judgments of multiple experts tailored to specific scenarios. Furthermore, each expert assesses candidate-job pairs from different perspectives, akin to the diverse evaluations by human professionals.



\section{Related Work}
In recent years, the PJF task has been regarded as a text matching task, aiming to fully utilize the rich semantic information in resumes and job descriptions. Various neural networks, such as CNN~\cite{zhu2018person,maheshwary2018matching,he2021finn,zhenhong2021person}, RNN~\cite{qin2018enhancing,qin2020enhanced,yan2019interview} and GNN~\cite{wang2022person} have been employed to encode resumes and JDs. However, these methods frequently overlook candidate-job interactions.

\paragraph{LLM-based methods} CONFIT~\cite{yu2024confit} uses LLMs to increase training samples by paraphrasing specific sections in resumes or job posts. In contrast, our method enhances low-quality job descriptions by polishing and rewriting based on historically matched resumes, which is orthogonal to CONFIT. Another method LGIR~\cite{du2024enhancing} improves job recommendations by combining explicit and implicit user data for better resume completion using LLMs. MockLLM~\cite{sun2024facilitating} introduces a mock interview and two-sided evaluation method, using an LLM agent to simulate both interviewer and candidate roles. Our method is also orthogonal to them, as LGIR focuses on resume completion, while MockLLM uses LLMs for mock interviews. A detailed literature review is provided in Appendix~\ref{appendix:rw}.





\section{Conclusion}
This paper presents innovative solutions to the persistent challenges in the Person-Job Fit (PJF) task, specifically addressing the issues of low-quality job descriptions and similar candidate-job pairs. By introducing an LLM-based method based on a data augmentation strategy, alongside the fine-grained historical interaction module and the category-aware MoE module, our approach significantly enhances the person-job matching process. Results from both offline experiments and online A/B tests demonstrate the superiority of our method. 

\section*{Limitations}
Our system depends on the availability of historical interaction sequences, which may be sparse for new users or job posts, potentially affecting matching accuracy in cold-start scenarios.  In practice, we can leverage LLM-generated pseudo-histories to improve robustness when interaction data is sparse. We plan to address this issue by developing a domain-specific LLM that can provide pre-assessments for the candidates and simulate the interview process to accumulate historical data. Additionally, our LLM-based data augmentation strategy relies on a fixed-length threshold to identify low-quality job descriptions, which may not fully capture the nuanced variability in content richness across different industries or job categories. 

\section*{Ethical considerations}
In conducting our research, we have prioritized ethical considerations to ensure the responsible use of data throughout the study. Our collected real-world data mainly consists of job descriptions and resumes of talents. To protect the privacy of users, we only use the professional experiences in their resumes as inputs for our person-job fit model and remove personally identifiable information. We also mitigate any potential biases in the dataset. As introduced in Section~\ref{strategies_for_bias}, sensitive features that may lead to unfairness and biases, e.g., gender and age, are excluded. Additionally, our system is equipped with human oversight and feedback mechanisms, ensuring that any potential bias can be promptly addressed.

\section*{Acknowledgments}
This work was supported by the Key R\&D Program of Zhejiang (2024C01036). It was also funded by the NSFC Project (No. 62306256) and the Natural Science Foundation of Guangdong Province (No. 2025A1515010261).



\bibliography{main}

\appendix
\section{COT Prompt Template}
\label{prompt_example}

\begin{figure}[h]
\begin{center}
\includegraphics[width=\linewidth]{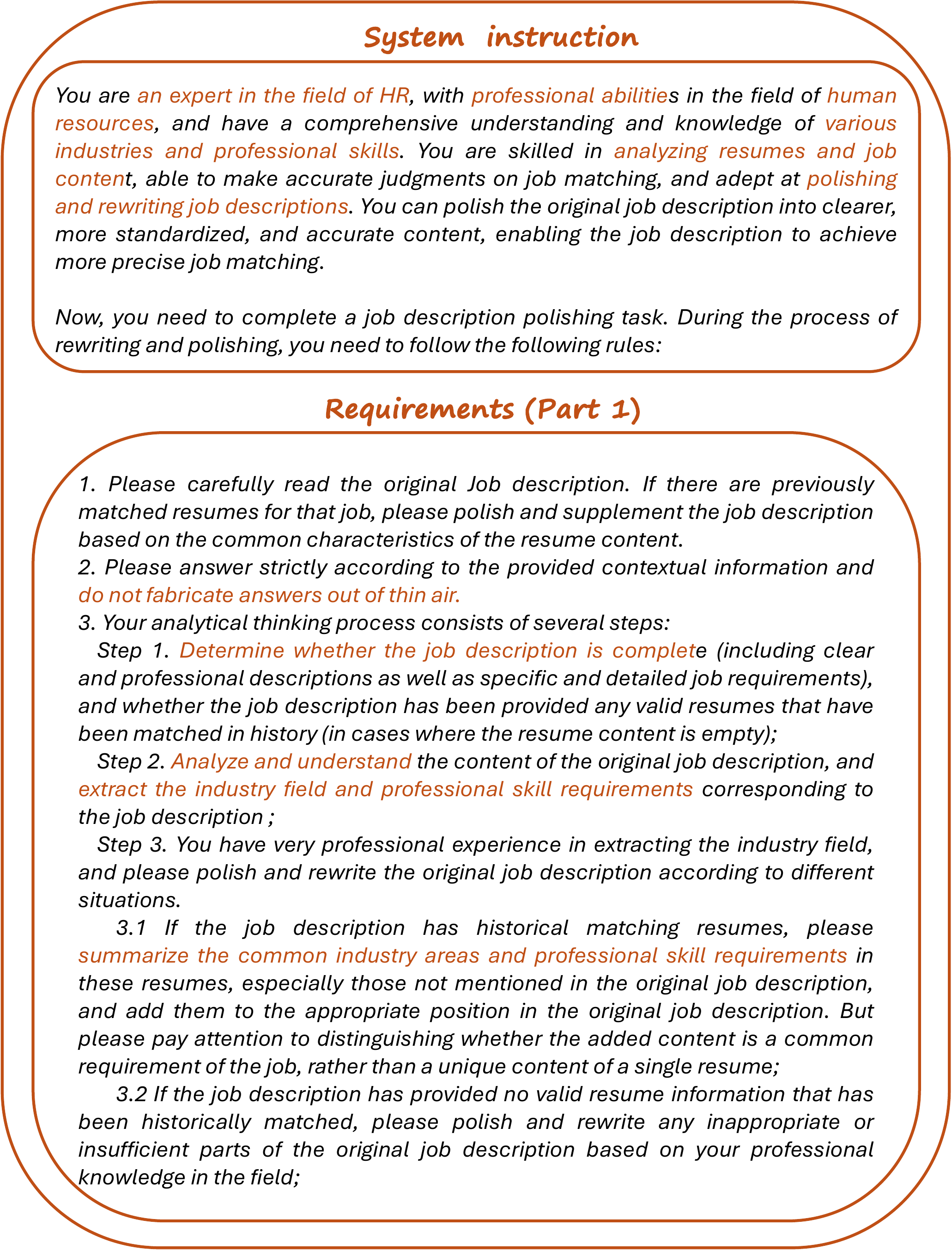}
\end{center}
\caption{An example of the designed prompt (Part 1).} 
\label{prompt-1}
\end{figure}

\begin{figure}[h]
\begin{center}
\includegraphics[width=\linewidth]{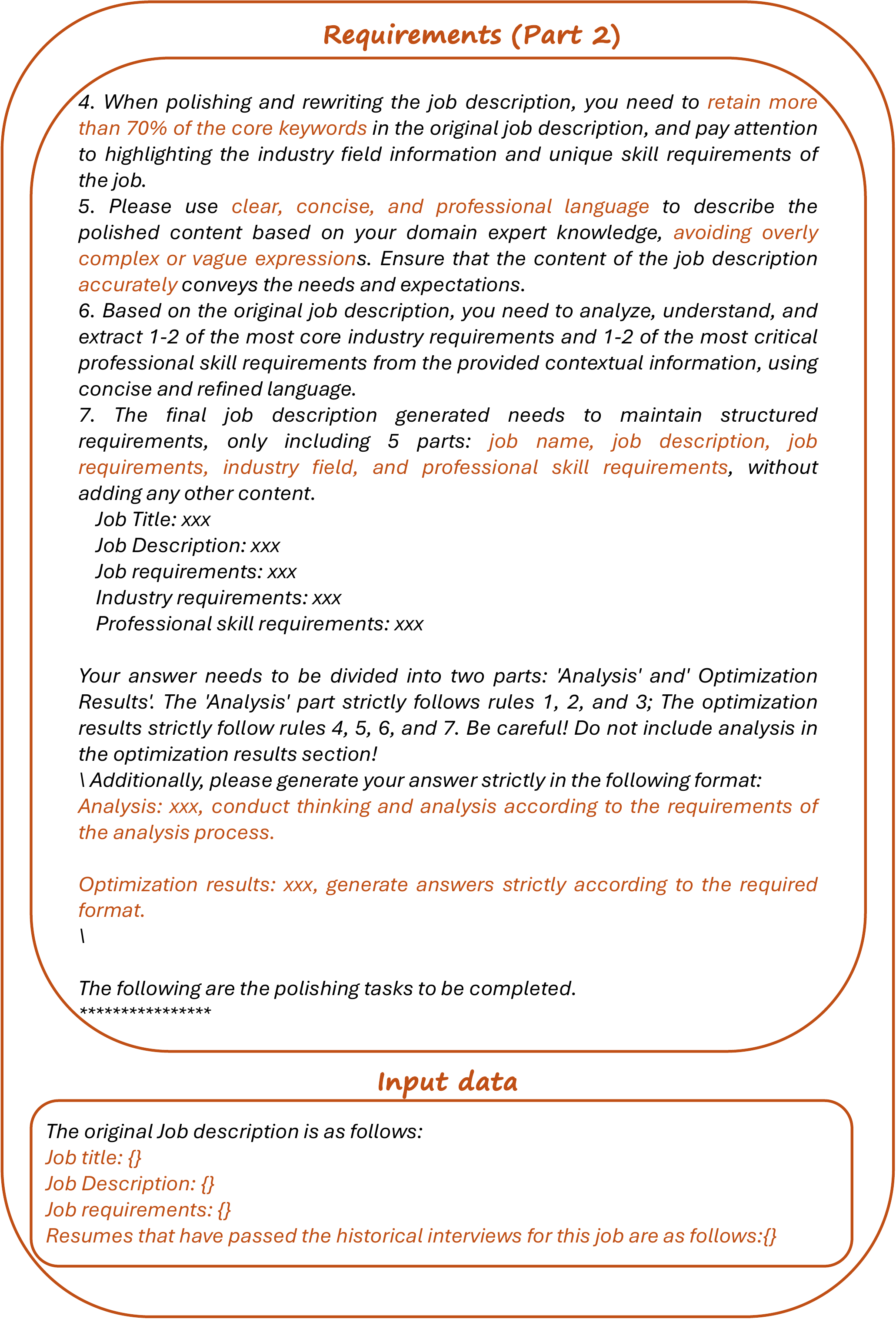}
\end{center}
\caption{An example of the designed prompt (Part 2).} 
\label{prompt-2}
\end{figure}

An example of the complete prompt utilized in our LLM-based data augmentation is illustrated in Figure~\ref{prompt-1} (Part 1) and Figure~\ref{prompt-2} (Part 2). Key elements within the prompt are emphasized to draw attention to their significance. We customize the prompt to adapt to different job categories for better data augmentation quality.

\section{System Workflow}
\begin{figure}[t]
\begin{center}
    \includegraphics[width=0.95\linewidth]{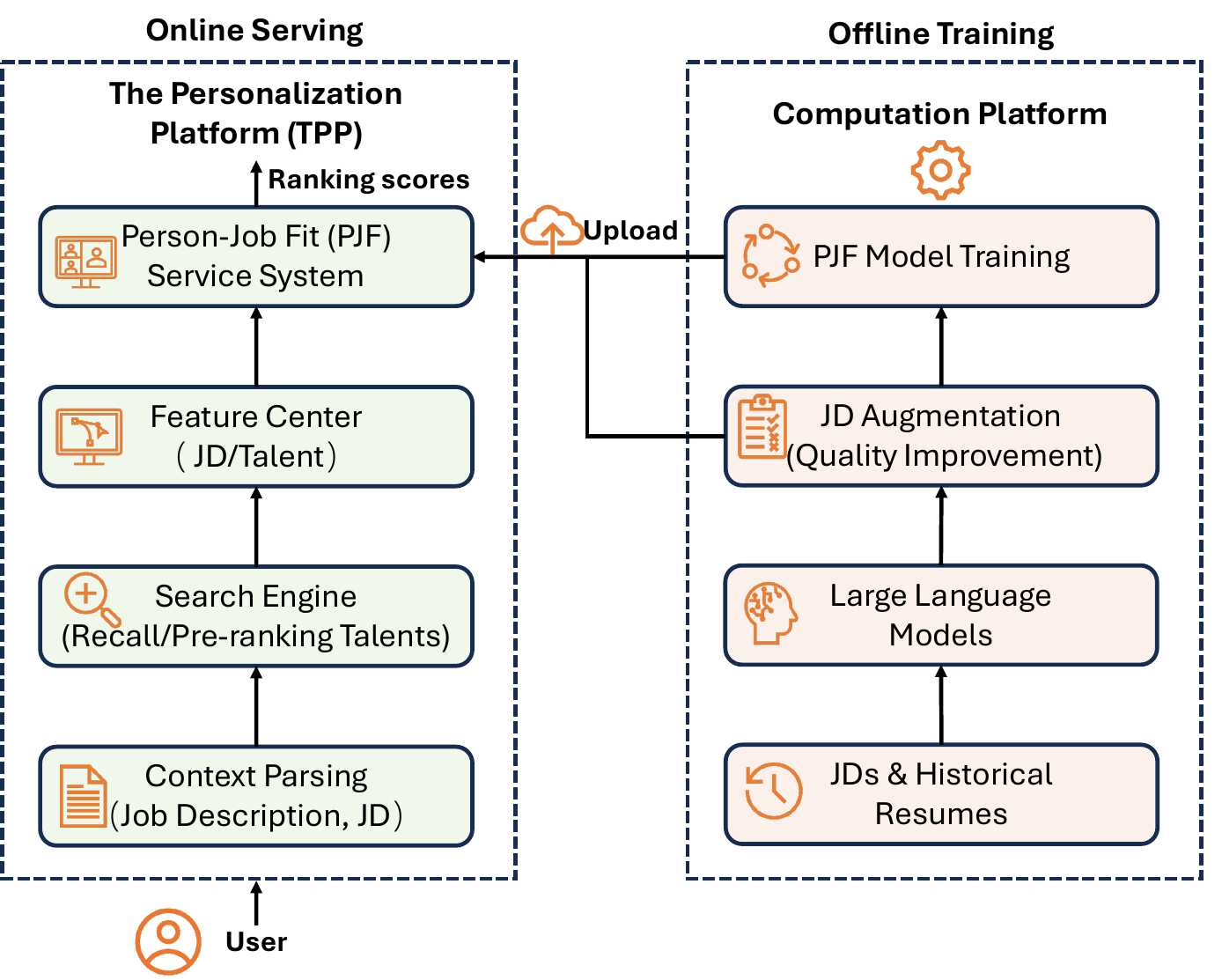}
\end{center}
\caption{Workflow of our online system.} 
\label{workflow}
\end{figure}

Here, we present a comprehensive overview of the workflow of our online system. As illustrated in Figure~\ref{workflow}, our system is architected with two primary components: the online serving module and the offline training module. The online serving module performs essential pre-processing, including parsing the real-time job description (JD) provided by the user, recalling and pre-ranking talents via a search engine, and constructing features for both the JDs and talents. These features are subsequently input to the PJF service system to compute the ranking scores for the candidates. The offline training module is responsible for optimizing the PJF model, which is enhanced with the LLM-based JD augmentation module by leveraging the historical resumes. The refined PJF model, along with the augmented JDs, is then uploaded to the PJF service system for online inference.

\section{Experimental Setup}
\label{setup}
\sloppy This section introduces the key implementation and baseline details. We divide a small subset from the training data for hyperparameter tuning, and the test set is only used for final evaluation. 

\paragraph{Embedding} We use the fine-tuned BGE-M3 model~\cite{chen2024bge} to convert the input texts, i.e., resumes and job descriptions, into 1024-dimensional embeddings. The category vocabulary size $n_{\rm c}$ is 16, and the category embedding dimension $d_{\rm e}$ is 8. Specifically, we have the following categories: Technology, Product, Supply Chain, Logistics, Content, Customer Experience, Marketing, Advertising, Data, Gaming, General, Design, Operations, Sales, Project management, and Risk management. The text embedding matrix is frozen, while the category embedding matrix is trainable. 

\paragraph{LLM for Data Augmentation} 
We use Qwen1.5-72B-Chat\footnote{https://qwenlm.github.io/blog/qwen1.5/} to perform data augmentation for the low-quality job descriptions. We find that the quality of augmented job descriptions generated by Qwen1.5-72B-Chat is comparable to those produced by Qwen2.5-72B and Qwen3, with no significant differences observed. Our experimental results also demonstrate that Qwen1.5-72B-Chat is strong enough for job description augmentation.

\paragraph{Model Configuration} 
The character-level length threshold $l$ for LLM-based job description augmentation is 200. The category-aware MoE module has five experts, and the historical interaction sequence length is set to 20, with a padding operation for those historical interactions shorter than 20. In the multi-head attention, $d_{\rm model}$ is 1024 and $h$ is 2. 

\paragraph{Details of Evaluation Metrics}
The definitions of GAUC, normalized discounted cumulative gain (NDCG), average precision (AP), and click-through conversion rate (CTCVR) are as follows:

(1) GAUC:
\begin{equation*}
GAUC=\sum_{i=1}^{N_{\rm J}}\frac{N^{\rm J}_{i}}{N_{\rm t}}{AUC}_i,
\end{equation*}
where $N_{\rm J}$ is the number of jobs in the test set, $N_{i}^{\rm J}$ is the number of test samples associated with the $i$-th job, and $N_{\rm t}$ is the total number of test samples.

(2) NDCG:
\begin{equation*}
DCG=\sum_{i=1}^{N_{\rm t}}\frac{\hat{y}_i}{log_2(i+1)}, \\
\end{equation*}
\begin{equation*}
IDCG=\sum_{i=1}^{N_{\rm t}}\frac{\check{y}_i}{log_2(i+1)},
\end{equation*}
\begin{equation*}
NDCG=\frac{DCG}{IDCG},
\end{equation*}
where $\hat{y}_i$ and $\check{y}_i$ are the ground truths of the $i$-th test sample, where the test samples are sorted in descending order according to the corresponding predicted scores and the ground truths, respectively.

(3) AP: 
\begin{equation*}
AP =\sum_{i=0}^{N_{\rm th}-1}(recall_{i} - recall_{i+1})\times precision_i,
\end{equation*}
where $N_{\rm th}$ is the number of thresholds to transform a prediction score into a binary class, $recall_i$ and $precision_i$ are the recall and precision on the test set at the $i$-th threshold.

(4) CTCVR:
\begin{equation*}
CTR=\frac{N_{\rm click}}{N_{\rm PV}}, CVR=\frac{N_{\rm application}}{N_{\rm click}}
\end{equation*}
\begin{equation*}
CTCVR=CTR*CVR=\frac{N_{\rm application}}{N_{\rm PV}},
\end{equation*}
where $N_{\rm click}$ is the total number of clicks, $N_{\rm PV}$ is the total search exposure, and $N_{\rm application}$ is the number of application initiations in our system. A higher CTCVR value indicates a more streamlined process for recruiters, enabling them to locate ideal candidates with greater accuracy and reduced effort, thereby minimizing the number of resumes they need to review. 

\paragraph{Training Setup} 
The training batch size, learning rate, and the weight of the regularization item are set to 256, 1e-4, and 0.1, respectively. Given our large training dataset (about 8 million), the model is sufficiently trained after one epoch, so we evaluate the test set after one epoch. 

\paragraph{Baseline Details} We perform hyperparameter tuning for all baselines to obtain their optimal performance on our dataset. Here, we provide a brief introduction of each baseline in our experiments:
\begin{enumerate}
\item \textbf{Logistic Regression (LR)}/\textbf{XGBoost}~\cite{chen2016xgboost}: The text embeddings in our method are used as features for prediction.
\item\textbf{DSSM}~\cite{huang2013learning}: It projects the resumes and the job descriptions into a common low-dimensional space and uses cosine similarity to determine their matching degree. 
\item \textbf{BGE-Raw}~\cite{chen2024bge}: It uses the BGE-M3 model to produce the embeddings of the resumes and the job descriptions, with matching determined by cosine similarity.
\item \textbf{PJFNN}~\cite{zhu2018person}: It encodes the resumes and job descriptions independently by a hierarchical CNN, and the matching degree is also determined by cosine similarity.
\item \textbf{IPJF}~\cite{le2019towards}: It utilizes multiple labels to represent the propensity of candidates and jobs forming a match. 
\item \textbf{PJFFF}~\cite{jiang2020learning}: It learns implicit intentions of the candidates or recruiters from historical accepted and rejected applications. 
\item \textbf{SHPJF}~\cite{hou2022leveraging}: It utilizes both text content from
jobs/resumes and search histories from users.
\item \textbf{CONFIT}~\cite{yu2024confit}: It increases the number of training samples using LLM-based data augmentation techniques, and exploits contrastive learning to train a transformer-based encoder.
\end{enumerate}

\begin{figure}[t]
\begin{minipage}[h]{0.48\linewidth}
\centering
\subfloat[Number of experts]{
\includegraphics[width=4cm]{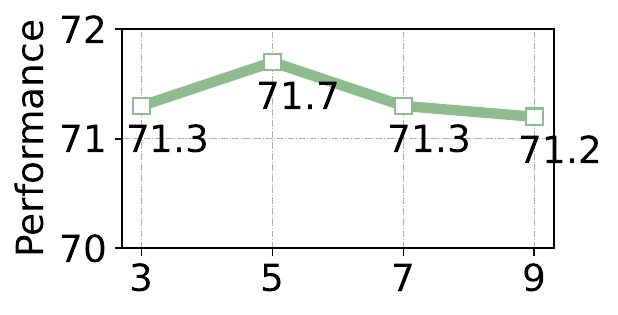}}
\end{minipage}
\begin{minipage}[h]{0.48\linewidth}
\centering
\subfloat[Interaction sequence length]{
\includegraphics[width=4cm]{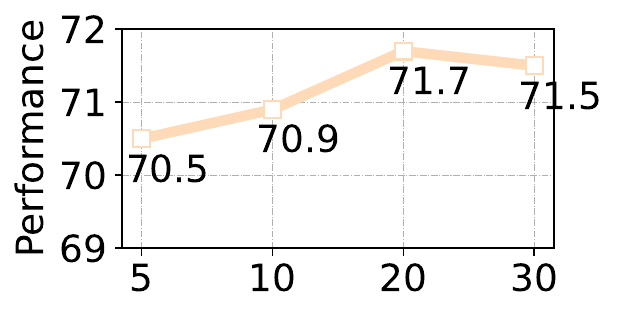}}
\end{minipage}

\caption{Impact of two hyperparameters on the final performance (average result of AUC and GAUC).}
\label{impact_exp}
\end{figure}

\begin{figure*}[t]
\begin{center}
    \includegraphics[width=0.95\linewidth]{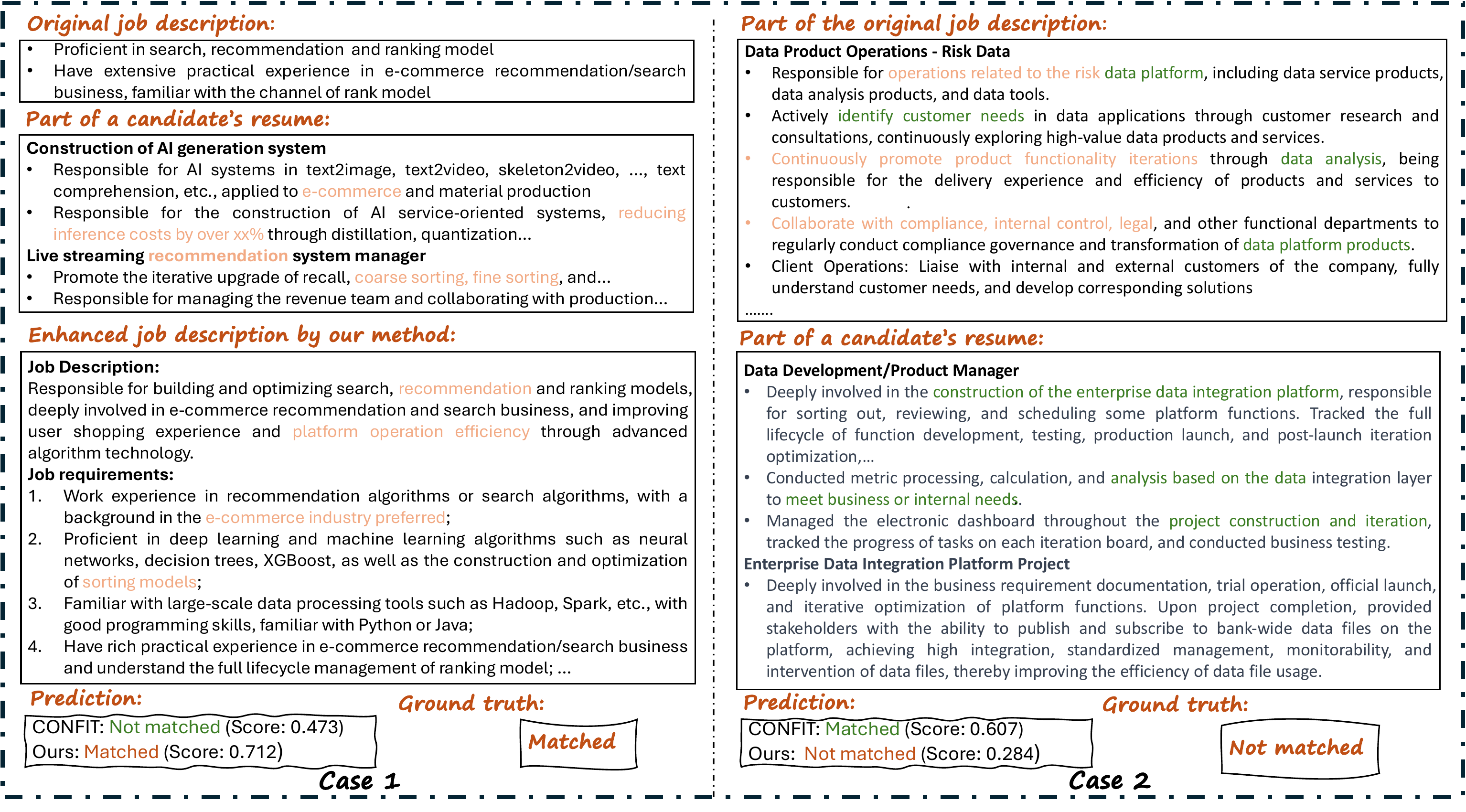}
\end{center}
\caption{Case study on LLM-based job description augmentation and similar candidate-job pair.} 
\label{case_study_1}
\end{figure*}

\section{Extended Related Work}
\label{appendix:rw}
The Person-Job Fit (PJF) task has received extensive scholarly attention, and it originated from the work of~\citet{malinowski2006matching}, which uses the expectation maximization algorithm based on candidate and job opportunity profiles. Traditional studies have approached this task using collaborative filtering~\cite{diaby2013toward,lu2013recommender,zhang2014research}, while neural networks have been extensively used for this task in recent years~\cite{qin2020enhanced,wang2022person}.

\paragraph{Modeling interactions} Some existing methods explore various interaction strategies between candidates and jobs to improve the performance of the PJF model~\cite{le2019towards,fu2021beyond,yang2022modeling,hou2022leveraging,yao2022knowledge,su2022optimizing}. For instance, \citet{zheng2023reciprocal} formulate reciprocal recommendation as a distinctive sequence matching task and propose to leverage bilateral behavior sequences for dynamic interest modeling on both sides. \citet{zheng2024mirror} propose a MultI-view Reciprocal Recommender system (MIRROR) which models the users from three different views to capture the user representation corresponding to each view. Compared with the existing methods, the interactions between candidates and jobs of our method are more fine-grained and comprehensive.

\section{More Experiments}
\subsection{Impact of Hyperparameters} 
The ablation study in Table~\ref{ablation_study} highlights the importance of the category-aware MoE module. Here, we investigate the impact of the number of experts in the MoE module. Results in Figure~\ref{impact_exp} (a) indicate that five experts yield optimal performance, while increasing the number to seven or nine experts results in performance decline, possibly due to the over-fitting issue. We also carry out experiments to examine how the length of historical interaction sequences affects performance. The findings, illustrated in Figure~\ref{impact_exp} (b), indicate that performance improves as the sequence length increases up to 20. Beyond this point, longer sequences do not enhance performance and reduce training efficiency. Thus, we set the interaction sequence length to 20 in our experiments.

\subsection{More Detailed Ablation Study}
Here, we conduct a more detailed ablation study of our category-aware MoE module, as shown in Table~\ref{moe_ablation_study}. \textit{w/o MoE} refers to the model that disables the whole MoE module. \textit{MoE-category} and \textit{MoE-experts} refer to the model removing the category embedding and the experts in the MoE module, while other components in this module remain the same. \textit{Ours (simple match)} is a simple alternative to the MoE module that uses the binary match score (0/1) between the job category and the candidate category to concatenate with other representations. The results of removing the category embedding and experts alone suggest that they both contribute to performance improvements. Additionally, the results of using a simple match to replace the MoE module demonstrate that our MoE module is essential for the PJF task.
\begin{table}[t]
\centering
\caption{Ablation study of the category-aware MoE module in our method.} 
\resizebox{0.85\linewidth}{!}{
\newcommand{\tabincell}[2]{\begin{tabular}{@{}#1@{}}#2\end{tabular}}
\begin{tabular}{l|cc}
\toprule
Method & AUC &  GAUC  \\
\midrule
\midrule
Ours (full model) & \textbf{0.724}  & \textbf{0.709} \\
w/o MoE & 0.679 & 0.665 \\
w/o MoE-category & 0.698 & 0.681 \\
w/o MoE-experts &   0.684 & 0.667  \\
Ours (simple match) &   0.694 & 0.680 \\
\bottomrule
\end{tabular} 
}
\label{moe_ablation_study}
\end{table}

\subsection{Case Study}
Here, we present two illustrative cases to demonstrate how our method outperforms the SOTA baseline CONFIT. As shown in Figure~\ref{case_study_1}, \textit{Case 1} shows a low-quality original job description (JD) with limited content and information. CONFIT fails to predict this case accurately (the threshold for transforming the matching score to a binary label is 0.5). In comparison, our method enhances the JD by refining the content and adding some common requirements based on the provided historical resumes and the strong LLM knowledge, leading to accurate predictions. In \textit{Case 2}, the JD and the candidate's resume exhibit several similarities. However, the resume emphasizes data development and product management in enterprise data integration, lacking the operational focus on risk data and customer interaction required by the JD. Additionally, it does not show collaboration with compliance and governance teams, which are key aspects of the job. With the assistance of the category-aware MoE module, our method accurately predicts this case, while CONFIT predicts the wrong label.


\end{document}